\documentclass[letterpaper, 10 pt, conference]{ieeeconf}  

\IEEEoverridecommandlockouts                              

\usepackage{graphics} 
\usepackage{epsfig} 
\usepackage{mathptmx} 
\usepackage{times} 
\usepackage{amsmath} 
\usepackage{amssymb}  
\usepackage{url}
\usepackage{booktabs}
\usepackage{siunitx}
\usepackage{csquotes}
\usepackage{pgf}
\usepackage{pgfplots}
\usepackage[T1]{fontenc}
\usepackage{tikz}
\usepackage{makecell}

\usepackage{multirow} 
\usepackage{pifont} 
\usepackage{arydshln} 
\usepackage{subfiles}
\usepackage{balance} 
\usepackage[absolute]{textpos} 
\usepackage{xcolor}
\usepackage{dsfont}

\usepackage[caption=false, font=footnotesize]{subfig}

\let\labelindent\relax
\usepackage{enumitem}

\newcolumntype{x}[1]{>{\centering\arraybackslash\hspace{0pt}}p{#1}}

\usepackage{caption}
\usepackage{etoolbox}
\makeatletter
\patchcmd{\@makecaption}
  {\scshape}
  {}
  {}
  {}
\makeatletter
\patchcmd{\@makecaption}
  {\\}
  {.\ }
  {}
  {}
\let\NAT@parse\undefined
\makeatother

\title{\LARGE \bf
Excavating in the Wild:\\The GOOSE-Ex Dataset for Semantic Segmentation
}

\usepackage[backref=page]{hyperref}
\hypersetup{colorlinks=true, 
	unicode=true, 
	linkcolor=[rgb]{0.10,0.05,0.67}, 
	citecolor=green, 
	filecolor=[rgb]{0.10,0.05,0.67}, 
	urlcolor=[RGB]{237, 15, 152} 
}

\let\oldtwocolumn\twocolumn
\renewcommand\twocolumn[1][]{%
    \oldtwocolumn[{#1}{
    \begin{center}
           \includegraphics[width=\linewidth]{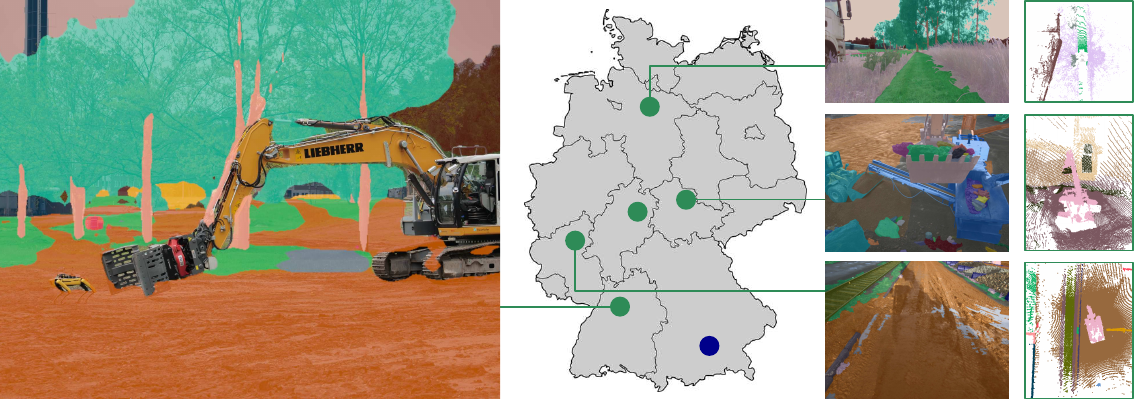}\\[12pt]
            \begin{minipage}{\textwidth}
              \raggedright
           \scriptsize \hypertarget{fig1}{Fig 1}.~~The GOOSE-Ex dataset was recorded over the course of a year in various locations across Germany, covering a wide range of environmental conditions. The left image shows a smartphone image of the two recording platforms, which was semantically segmented using a model trained on a 64-class version of the GOOSE-Ex dataset to demonstrate the platform generalizability of the dataset. The masks of the classes \textit{heavy\_machinery} and \textit{obstacle} were removed to better highlight the platforms. The right part of the figure shows exemplary ground truth annotations from different recording locations. This includes both pixel annotated images and annotated point clouds.
           \end{minipage}
          \label{fig:teaser}
        \end{center}
    }]
}

\newcommand{\copyrightstatement}{
	\begin{textblock*}{17cm}(20mm,1mm)    
		\noindent
		\footnotesize
		\copyright 2025 IEEE. Personal use of this material is permitted. Permission from IEEE must be
		obtained for all other uses, in any current or future media, including
		reprinting/republishing this material for advertising or promotional purposes, creating new
		collective works, for resale or redistribution to servers or lists, or reuse of any copyrighted
		component of this work in other works.
	\end{textblock*}
}

\newcommand{\underReviewStatement}{
	\begin{textblock*}{17cm}(20mm,1mm)    
		\noindent
		\footnotesize
		This is a preprint of a paper submitted to an IEEE journal/conference. 
     The paper is currently under review. 
     Please cite the final published version when available.
	\end{textblock*}
}

\newif\ifaddUnderReviewNotice
\addUnderReviewNoticefalse

\newif\ifaddIEEECopyrightNotice
\addIEEECopyrightNoticetrue

\author{Raphael Hagmanns$^{1,3}$, Peter Mortimer$^{2}$, Miguel Granero$^{1}$, Thorsten Luettel$^{2}$ and Janko Petereit$^{1}$
\thanks{This work was supported by the Federal Office of Bundeswehr Equipment, Information Technology and In-Service Support (BAAINBw).}%
\thanks{$^{1}$ The authors are with the Fraunhofer Institute of Optronics, System Technologies and Image Exploitation, Karlsruhe, Germany {\tt\small firstname.lastname@iosb.fraunhofer.de}}%
\thanks{$^{2}$ The authors are with the Institute for Autonomous Systems Technology, University of the Bundeswehr Munich, Germany
        {\tt\small firstname.lastname@unibw.de}}%
\thanks{$^{3}$ The author is with the Karlsruhe Institute of Technology (KIT), Germany
        {\tt\small firstname.lastname@kit.edu}}%
}

\usepackage[numbers,sort&compress]{natbib}

\begin{document}

\ifaddUnderReviewNotice
	\underReviewStatement
\fi

\ifaddIEEECopyrightNotice
	\copyrightstatement
\fi

\setcounter{figure}{1}

\definecolor{windshield}{RGB}{0, 170, 0}
\definecolor{os128}{RGB}{0, 255, 255}
\definecolor{os64}{RGB}{197, 141, 83}
\definecolor{os64rev7}{RGB}{255, 165, 0}
\definecolor{surround}{RGB}{0, 0, 255}
\definecolor{surroundSpot}{RGB}{170, 0, 0}
\definecolor{ellipse}{RGB}{255, 85, 127}
\definecolor{ekinox}{RGB}{170, 0, 255}
\definecolor{realsense}{RGB}{236, 146, 56}

\definecolor{undefined}{RGB}{0,0,0}
\definecolor{traffic_cone}{RGB}{255,255,0}
\definecolor{snow}{RGB}{209,87,160}
\definecolor{cobble}{RGB}{255,52,255}
\definecolor{obstacle}{RGB}{255,74,70}
\definecolor{leaves}{RGB}{0,137,65}
\definecolor{street_light}{RGB}{0,111,166}
\definecolor{bikeway}{RGB}{163,0,89}
\definecolor{ego_vehicle}{RGB}{255,219,229}
\definecolor{pedestrian_crossing}{RGB}{122,73,0}
\definecolor{road_block}{RGB}{0,0,166}
\definecolor{road_marking}{RGB}{99,255,172}
\definecolor{car}{RGB}{183,151,98}
\definecolor{bicycle}{RGB}{0,77,67}
\definecolor{person}{RGB}{143,176,255}
\definecolor{bus}{RGB}{153,125,135}
\definecolor{forest}{RGB}{90,0,7}
\definecolor{bush}{RGB}{128,150,147}
\definecolor{traffic_light}{RGB}{27,68,0}
\definecolor{motorcycle}{RGB}{79,198,1}
\definecolor{sidewalk}{RGB}{59,93,255}
\definecolor{curb}{RGB}{74,59,83}
\definecolor{asphalt}{RGB}{255,47,128}
\definecolor{gravel}{RGB}{97,97,90}
\definecolor{boom_barrier}{RGB}{52,54,45}
\definecolor{rail_track}{RGB}{107,121,0}
\definecolor{tree_crown}{RGB}{0,194,160}
\definecolor{tree_root}{RGB}{196,164,132}
\definecolor{tree_trunk}{RGB}{255,170,146}
\definecolor{debris}{RGB}{136,111,76}
\definecolor{crops}{RGB}{0,134,237}
\definecolor{soil}{RGB}{209,97,0}
\definecolor{rider}{RGB}{221,239,255}
\definecolor{animal}{RGB}{0,0,53}
\definecolor{truck}{RGB}{123,79,75}
\definecolor{on_rails}{RGB}{161,194,153}
\definecolor{caravan}{RGB}{48,0,24}
\definecolor{trailer}{RGB}{10,166,216}
\definecolor{building}{RGB}{1,51,73}
\definecolor{wall}{RGB}{0,132,111}
\definecolor{rock}{RGB}{55,33,1}
\definecolor{fence}{RGB}{255,181,0}
\definecolor{guard_rail}{RGB}{194,255,237}
\definecolor{bridge}{RGB}{160,121,191}
\definecolor{tunnel}{RGB}{204,7,68}
\definecolor{pole}{RGB}{192,185,178}
\definecolor{traffic_sign}{RGB}{194,255,153}
\definecolor{misc_sign}{RGB}{0,30,9}
\definecolor{barrier_tape}{RGB}{190,196,89}
\definecolor{kick_scooter}{RGB}{111,0,98}
\definecolor{low_grass}{RGB}{12,189,102}
\definecolor{high_grass}{RGB}{238,195,255}
\definecolor{scenery_vegetation}{RGB}{69,109,117}
\definecolor{sky}{RGB}{183,123,104}
\definecolor{water}{RGB}{122,135,161}
\definecolor{wire}{RGB}{255,140,0}
\definecolor{outlier}{RGB}{120,141,102}
\definecolor{heavy_machinery}{RGB}{250,208,159}
\definecolor{container}{RGB}{255,138,154}
\definecolor{hedge}{RGB}{209, 87, 160}
\definecolor{moss}{RGB}{180, 168, 189}
\definecolor{barrel}{RGB}{208, 208, 0}
\definecolor{pipe}{RGB}{221, 0, 0}
\definecolor{military_vehicle}{RGB}{64,64,64}

\definecolor{vegetation}{RGB}{76, 175, 80}
\definecolor{terrain}{RGB}{161, 136, 127}
\definecolor{sky}{RGB}{33, 150, 243}
\definecolor{construction}{RGB}{255, 235, 59}
\definecolor{vehicle}{RGB}{244, 67, 54}
\definecolor{road}{RGB}{158, 158, 158}
\definecolor{object}{RGB}{255, 193, 7}
\definecolor{void}{RGB}{0, 0, 0}
\definecolor{other}{RGB}{169, 169, 169}

\maketitle
\thispagestyle{empty}
\pagestyle{empty}

\begin{abstract}
	The successful deployment of deep learning-based techniques for autonomous systems is highly dependent on the data availability for the respective system in its deployment environment. Especially for unstructured outdoor environments, very few datasets exist for even fewer robotic platforms and scenarios. In an earlier work, we presented the German Outdoor and Offroad Dataset (GOOSE) framework along with $\mathbf{10\,000}$ multimodal frames from an offroad vehicle to enhance the perception capabilities in unstructured environments. In this work, we address the generalizability of the GOOSE framework.  To accomplish this, we open-source the GOOSE-Ex dataset, which contains additional $\mathbf{5\,000}$ labeled multimodal frames from various completely different environments, recorded on a robotic excavator and a quadruped platform. We perform a comprehensive analysis of the semantic segmentation performance on different platforms and sensor modalities in unseen environments. In addition, we demonstrate how the combined datasets can be utilized for different downstream applications or competitions such as offroad navigation, object manipulation or scene completion. The dataset, its platform documentation and pre-trained state-of-the-art models for offroad perception will be made available on \url{https://goose-dataset.de/}.
	\\

\end{abstract}

\vspace{1em} 
The perception of unstructured outdoor environments presents significant challenges for autonomous systems, particularly in the domains of free space detection and obstacle avoidance, as well as for manipulation tasks. Attaining complete autonomy in these settings is challenging due to the inherent variability of environmental conditions and terrain types. In recent years, some effort has been made to adapt advanced semantic segmentation models from structured to unstructured environments. However, the adaptation of these models to previously unseen environments and new platforms is a challenging task due to the limited data availability. A particular challenge arises from the platform gap resulting from the platform-specific mounting of cameras and LiDAR sensors, which complicates the transfer-learning to different systems. The GOOSE framework presented in~\cite{mortimer_goose_2024} offers a robust basis for segmentation tasks, yet it is constrained to a single platform and a relatively small region. The main objective of the proposed GOOSE-Ex dataset is to facilitate adaptation to heterogeneous platform settings in specialized environments. Platform variations include an excavator setup as well as two quadruped robot setups with varying sensors. We also add out-of-distribution sequences to enable robustness investigations.

This paper presents a series of contributions designed to enhance the perception of various robots in diverse unstructured environments.

\begin{table*}[htpb!]
	\vspace{1em}
	\centering
	\caption{
		Comparison of sizes and sensor modalities between existing offroad datasets.
		In terms of size, the RELLIS-3D and GOOSE provides more annotated laser scans, but fewer annotated images.
		The CWT dataset also contains annotated sensor data from an autonomous excavator, but is notably smaller than GOOSE-Ex.
		For the parts of GOOSE-Ex recorded with a quadrupedal robot platform, the RUGD and RELLIS-3D are the most similar datasets in terms of the camera and LiDAR sensor height above ground.
	}
	\renewcommand{\arraystretch}{1.1} 
	\begin{tabular}{@{}ccccccl@{}}
		\toprule
		Dataset                               & Platform        & Sensors                     & Annotated Sensor Modalities & \# Annotations    & \# Classes \\ \midrule
		CWT \cite{guan_tns_2022}              & Excavator       & camera                      & RGB                         & 669               & 7          \\
		RUGD \cite{wigness_rugd_2019}         & Husky           & camera                      & RGB                         & 7\,546            & 24         \\
		RELLIS-3D \cite{jiang_rellis-3d_2021} & Warthog         & stereo camera / LiDAR / INS & RGB+Depth / Point Cloud     & 6\,235 / 13\,556  & 20         \\
		GOOSE \cite{mortimer_goose_2024}      & MuCAR-3         & prism camera / LiDAR / INS  & RGB+NIR / Point Cloud       & 10\,000 / 10\,000 & 64         \\
		\hdashline
		GOOSE-Ex (\textbf{ours})              & Excavator, Spot & prism camera / LiDAR / INS  & RGB+(NIR) / Point Cloud     & 5\,000 / 5\,000   & 64         \\

		\bottomrule
	\end{tabular}
	\label{tab:datasets}
\end{table*}
\begin{itemize}
	\item We present the GOOSE-Ex dataset, which consists of 5\,000 calibrated pairs of pixel-wise annotated RGB images and point-wise annotated LiDAR point clouds from a robotic excavator and a quadruped platform. The dataset encompasses over 100 sequences from diverse environments, employing the same dataset format and class hierarchy established in the GOOSE framework~\cite{mortimer_goose_2024}.
	\item We open-source the dataset and accompanying tools to enable rapid prototyping. We also provide additional sensor data, such as near-infrared (NIR) channels of many camera frames, surround views, and a high-precision localization.
	\item We evaluate the performance of various state-of-the-art models for semantic segmentation across different dataset combinations and sensor modalities.
	\item To the best of our knowledge, GOOSE-Ex is the first large-scale semantic segmentation dataset for excavator platforms. This can accelerate the progress in a variety of downstream applications, of which we showcase some in Section~\ref{sec:downstream}.
\end{itemize}

\section{Related Work}
\label{sec:related-work}

The release of datasets with dense semantic and instance-wise annotations of pixels in color images~\cite{geiger_vision_2013,cordts_cityscapes_2016,yu_bdd100k_2020,zendel_wilddash2_2022} and 3D points in LiDAR scans~\cite{behley_semantickitti_2019,caesar_nuscenes_2020,sun_waymo_2020} have led to ever-improving segmentation models for perception in autonomous driving in urban environments.
In recent years, there have been attempts to replicate the results for navigation in unstructured outdoor environments~\cite{valada_deepscene_2016, metzger_tas500_2020, maturana_ycor_2018, wigness_rugd_2019, neigel_offsed_2021, jiang_rellis-3d_2021, nunes_synphorest_2022, sharma_CaT_2022, mortimer_goose_2024}.

Table~\ref{tab:datasets} gives an overview of semantic segmentation datasets similar to GOOSE-Ex and their main characteristics.
Not included in the comparison in Table~\ref{tab:datasets} are datasets that have been annotated in unstructured environments for other specific tasks such as offroad free space detection~\cite{hoveidar_trailnet_2018,hosseinpoor_vale_2021,min_orfd_2022,sharma_CaT_2022}, place recognition~\cite{knights_wildplaces_2023}, learning offroad dynamic models~\cite{triest_tartandrive_2022,sivaprakasam_tartandrive2_2024,datar_vertiwheelers_2024} or end-to-end driving~\cite{tampuu_estoniadriving_2023}.

Currently, the recent GOOSE dataset~\cite{mortimer_goose_2024} and the RUGD dataset~\cite{wigness_rugd_2019} include the largest number of annotated images primarily focused on offroad scenes.
The high reflectivity of foliage in the near-infrared (NIR) spectrum~\cite{wood_woodeffect_1910} motivated the inclusion of this image modailty in datasets like TAS-NIR~\cite{mortimer_tasnir_2022}, Freiburg Forest~\cite{valada_deepscene_2016} and GOOSE~\cite{mortimer_goose_2024}.

Many of the early datasets in this domain like the OFFSED dataset~\cite{neigel_offsed_2021}, the TAS500 dataset~\cite{metzger_tas500_2020} and the YCOR dataset~\cite{maturana_ycor_2018} lack the size and variety to train deep neural networks that can generalize to a different robot platform.

Among the 3D point cloud datasets, the RELLIS-3D dataset~\cite{jiang_rellis-3d_2021} and the GOOSE dataset~\cite{mortimer_goose_2024} contain fused LiDAR point cloud data of each annotated scene.

Semantic segmentation datasets have extended on existing datasets before.
IDD dataset~\cite{varma_idd_2019} extended the semantic segmentation schema used in the CityScapes dataset~\cite{cordts_cityscapes_2016} to novel object classes and novel driving scenarios.
In a similar vein, datasets were extended with adverse weather and lighting conditions ~\cite{sakaridis_foggycityscapes_2018, sakaridis_foggycityscapes2_2018, hu_raincityscapes_2019, shaik_iddaw_2024, sakaridis_darkzurich_2019}.
GOOSE-Ex uses the same semantic segmentation scheme as the GOOSE dataset, but extends both the acquisition platforms and domains beyond those in GOOSE by including annotated data from an autonomous excavator and a quadruped robot platform.

A growing number of autonomous excavators~\cite{frese2022, guan_tns_2022} exist, but most methods focus on planning excavation tasks~\cite{wang_aesplanning_2021, guo_aestrajectory_2022, zhu_aesrl_2022}.
The recent CWT dataset~\cite{song_aes_2023} consists of 669 images annotated for objects and a few relevant terrain types (see Table~\ref{tab:datasets}).

Previous datasets recorded on quadrupedal robots have focused on robot navigation~\cite{eder_robonav_2022, karnan_scand_2022}, odometry~\cite{jung_coral_2024}, mapping~\cite{chaney_m3ed_2023} and 3D pose estimation~\cite{avogaro_harper_2024}.
Most image data from quadrupedal robots is unlabeled~\cite{angelini_habitatmonitoring_2023}, making RUGD~\cite{wigness_rugd_2019} and RELLIS-3D~\cite{jiang_rellis-3d_2021} the most similar semantic segmentation datasets in terms of the sensor height above ground~\cite{mortimer_outdoorsurvey_2024}.
For both platform types, GOOSE-Ex provides a novel contribution by providing semantically segmented camera and LiDAR data that allows for robustness and fine-tuning of the semantic segmentation models.

\section{The GOOSE-Ex Dataset}
\label{sec:goose-dataset}

We summarize the main aspects of the GOOSE framework in Section~\ref{subsec:framework}, which includes the organization of the dataset, its structure, ontology, metadata, labeling policy, and more. For additional details, we refer to our previous work~\cite{mortimer_goose_2024}, where we published these definitions along with the original GOOSE dataset. In the remaining sections, we discuss the GOOSE-Ex dataset in detail.

\subsection{GOOSE Framework}
\label{subsec:framework}

\begin{figure*}[htpb]
	\centering
	\resizebox{!}{.38\textheight}{\input{images/semantic_histogram/goose2d_semantic_histogram_no_group.pgf}}
	\resizebox{!}{.38\textheight}{\input{images/semantic_histogram/goose2d_pie_categories.pgf}}
	\caption{Best to inspect digitally. Histogram of the annotated pixels $\square$ and points \includegraphics[height=0.7em]{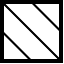}
		in the GOOSE-Ex dataset. The classes are split according to their pixel volume into high-, mid- and low-volume classes. We omitted the \textit{ego-vehicle} class in the point clouds and summarized classes with negligible occurrences in the \textit{other} bar.  The pie charts show the category distribution from the main categories \textit{Vegetation} \textcolor{vegetation}{$\blacksquare$}, \textit{Terrain} \textcolor{terrain}{$\blacksquare$}, \textit{Sky} \textcolor{sky}{$\blacksquare$}, \textit{Construction} \textcolor{construction}{$\blacksquare$}, \textit{Vehicle} \textcolor{vehicle}{$\blacksquare$}, \textit{Road} \textcolor{road}{$\blacksquare$}, \textit{Object} \textcolor{object}{$\blacksquare$}, \textit{Void} \textcolor{void}{$\blacksquare$}. The remaining categories \textit{Sign}, \textit{Human}, \textit{Water}, \textit{Animal} are summarized as \textit{Other} \textcolor{other}{$\blacksquare$}.\label{fig:histogram}}

\end{figure*}

Annotation of RGB images and LiDAR point clouds allows for 64 classes, enabling fine-grained segmentation tasks, especially for the traversability analysis across different vegetation and terrain types. It also enables fine-grained downstream applications such as the barrel detection presented in Section~\ref{sec:downstream}. For more general applications, we also suggest a rough division into categories, e.g. by aggregating all vegetation classes. The GOOSE ontology is inspired by different datasets and ontologies such as SemanticKITTI \cite{behley_semantickitti_2019}, TAS500 \cite{metzger_tas500_2020}, ATLAS~\cite{smith_atlas_2022}, and RELLIS-3D \cite{jiang_rellis-3d_2021} and designed to be as compatible and extendable as possible.

The GOOSE-Ex dataset was manually labeled, but multiple frames of a sequence were merged based on the platform odometry to facilitate the annotation process and increase the annotation quality. To achieve consistency across all datasets in the GOOSE framework, the same labeling policy was used for both the original GOOSE and GOOSE-Ex datasets. A hierarchical structure of the raw data allows easy filtering for environmental conditions or platform configurations. In addition to the raw data, we provide the labeled frames as a standalone data package in a format similar to the SemanticKITTI~\cite{behley_semantickitti_2019} dataset.

\subsection{Places}
\label{subsec:places}

Figure~\hyperlink{fig1}{1} shows the different recording areas of the GOOSE (blue) and GOOSE-Ex (green) datasets. We roughly divide the GOOSE-Ex dataset into four different high-level settings:

\begin{description}[]
	\item[generic] setting, containing a mixture of typical offroad and industrial regions
	\item[landfill] setting, containing frames from inside and around a landfill as a typical excavator operating environment
	\item[quarry] setting, as special operating environment for large machines, with complex surface geometries
	\item[construction site] setting, including an excavator training area with many different heavy machines
\end{description}

\noindent This subdivision allows to use specific parts of the dataset to fine-tuning different operational scenarios.

\subsection{Dataset Statistics}
\label{subsec:dataset-statistics}

The distribution of classes and categories is shown in Figure~\ref{fig:histogram}. For unstructured outdoor environments, the \textit{vegetation} and \textit{terrain} categories naturally make up the majority of the dataset. Typical for excavator scenarios, \textit{soil} is a dominant class in the distribution across all environments. The partitioning of the histogram into different settings reveals plausible effects: The \textit{generic} environment contains the most vegetation and many different mid-volume classes. Due to steep slopes in the terrain, the landfill setting contains more \textit{sky} than the others. The piles of trash at the landfill also account for the increase of \textit{debris} and \textit{obstacles}. \textit {Gravel} and \textit{rock} are naturally the most frequent classes in the quarry setting.
Other classes are very rare in the quarry environment, except \textit{heavy machinery} due to the large size of those vehicles in quarry environments. Finally, the construction site setting contains an equal appearance of classes with some outliers such as \textit{road blocks} and \textit{fences}.

Of the 64 classes present in the GOOSE ontology, only 36 are present in the histogram, all remaining occurrences are summarized in the \textit{other} bars. This illustrates the general problem of class imbalance in the natural environment. However, we believe that fine-grained annotations can only be advantageous as they allow to solve fine-grained object recognition and segmentation tasks. If this kind of granularity is not important for the task at hand, one can always refer to the coarser division into categories.

\section{Platform Setups}
\label{sec:robot-setup}
\begin{figure}[htpb!]
	\centering
	\includegraphics[width=1.0\columnwidth, page=1]{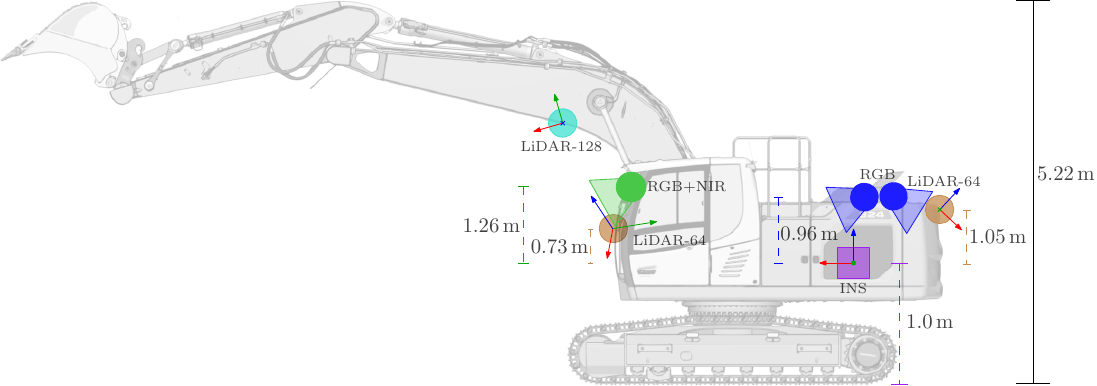} \\[12pt]
	\includegraphics[width=1.0\columnwidth, page=2]{images/sensor_setup/alice.pdf} \\[12pt]
	\includegraphics[width=1.0\columnwidth, page=3]{images/sensor_setup/alice.pdf}
	\caption{
		Schematic of the sensor setups on the Fraunhofer IOSB research excavator \textit{ALICE} and quadruped robot \textit{SpotLow}. All measurements are relative to the IMU frame.
	}
	\label{fig:sensor_setup}
\end{figure}

To enhance the transfer learning possibilities onto unique platforms, the GOOSE-Ex dataset was recorded on two robotic platforms as illustrated in Figure~\ref{fig:sensor_setup}.

\subsection{Sensor Setup}
\label{subsec:sensor-setup}

The main recording platform is the Liebherr R924 track excavator \textit{ALICE}~\cite{frese2022} with drive-by-wire capabilities and many modifications such as sensors for precise angle, force and track odometry measurements. The main sensors we use for the dataset collection include:

\begin{itemize}
	\item 3 $\times$ Ouster OS1-64 Rev6 (\textcolor{os64}{$\blacksquare$})~\cite{ousterDatasheetsFiles}
	\item 1 $\times$ Ouster OS1-128 Rev6 (\textcolor{os128}{$\blacksquare$}), mounted at boom
	\item 2 $\times$ JAI FSFE-3200D (\textcolor{windshield}{$\blacksquare$}): global shutter prism camera with two 3.2MP sensors for both RGB and NIR (near-infrared) equipped with a \SI{6}{\milli\meter} 1/1.8\textquotedbl~Fujinon TF6MA-1 lens, \SI{5}{\hertz}, \SI{59}{\degree} hFoV
	\item 4 $\times$ Alvium G1-240C (\textcolor{surround}{$\blacksquare$}): global shutter camera with 2.4MP resolution RGB sensor equipped with a Lensagon B5M3428S123C lens, \SI{5}{\hertz}, \SI{110}{\degree} hFoV
	\item SBG Ekinox-D (\textcolor{ekinox}{$\blacksquare$}): Inertial Navigation System (INS) with differential RTK-GNSS corrections received over a base-station if available or over LTE using the NTRIP protocol
\end{itemize}

As second platform, we equipped a Boston Dynamics Spot robot with two different sensor setups to further increase the platform generalizability. The setup includes computing hardware as well as a differential SBG Ellipse-D RTK-INS (\textcolor{ellipse}{$\blacksquare$}) for precise localization. The base platform already includes six low-resolution surround camera streams, which we utilize for the point cloud labeling and include in the raw data. The remaining setup differs in the following way:

\setlist[description]{font=\normalfont\itshape}
\begin{description}
	\item[SpotLow] equipped with
	      \begin{itemize}
		      \item Ouster OS1-64 Rev6 (\textcolor{os64}{$\blacksquare$})
		      \item Intel\textsuperscript{\textcopyright} RealSense\textsuperscript{TM} LiDAR Camera L515 (\textcolor{realsense}{$\blacksquare$}), \SI{5}{\hertz}, \SI{70}{\degree} hFoV, rolling shutter
	      \end{itemize}
	\item[SpotHigh] equipped with
	      \begin{itemize}
		      \item Ouster OS1-64 Rev7 (\textcolor{os64rev7}{$\blacksquare$})
		      \item Alvium G1-240C (\textcolor{surround}{$\blacksquare$}) (see above)
	      \end{itemize}
\end{description}

\subsection{Synchronization}
\label{subsec:synchronization}

We utilize the Precision Time Protocol (PTP, IEEE 1588~\cite{ieee_ptp}) support of the sensors to synchronize the Ouster LiDARs with the cameras and the INS system. Only the Realsense\textsuperscript{TM} L515 camera of the \textit{SpotLow} does not support PTP, so we use the system clock with some exposure offset instead. During post-processing, we leverage the ROS inbuilt \textit{approximate time synchronizing} mechanism to match point clouds and camera images. The INS system provides the \textit{grandmaster} clock in the sensor network, receiving its time stamps with \SI{200}{\hertz} via GNSS.

\subsection{Calibration}
\label{subsec:calibration}

We leverage a custom calibration suite to determine both intrinsic and extrinsic parameters of all cameras and LiDARs. For intrinsic and stereo calibration, we assume the pinhole projection model and use a standard checkerboard calibration procedure within our suite. For extrinsic calibration, we build on~\cite{guindel_lidarcalibrationboard_2017} and~\cite{beltran_lidarcameracalibration_2022}, and use a calibration target with Apriltags and three circular holes to allow for target matching in both the point cloud and the camera image. One of our LiDAR scanners on the excavator moves along with the boom, so the extrinsic transformation would change as the excavator moves. We therefore merge all point clouds from different scanners with respect to the INS frame, resulting in a single consistent transformation that can be used for reprojection.
\\

\section{Experimental Evaluation}
\label{sec:evaluation}
\subsection{Training Split}
\label{subsec:training-split}
Similar to~\cite{mortimer_goose_2024}, we divided the dataset into \textit{scenarios} consisting of multiple \textit{sequences}, one per recorded rosbag. We select sequences from different scenarios to define training (3989 frames), validation (407 frames) and test (604 frames) splits. We withhold the label files for the small test split to include it in a public benchmark of all GOOSE datasets. Of the 5000 total frames, 2800 are from the excavator, the remaining 2200 from the Spot robot.

\begin{table*}[htpb!]
	\vspace{3pt}
	\centering
	\caption{Comparison of the 2D image segmentation and 3D point cloud segmentation performance on the GOOSE-Ex test set. The IoU scores are specified in percent. For class-based evaluation, classes with occurences less than 20 are omitted. No classes of the category \textit{Sky} exist for the 3D point cloud segmentation. The models were trained using the GOOSE dataset as a base and fine-tuned on the GOOSE-Ex dataset. \textit{Category} models are trained directly on category labels, whereas \textit{class} IoU values are calculated per class and averaged afterwards.}
	\renewcommand{\arraystretch}{1.1}
	\begin{tabular}{*{8}{cllcccccccccc}}
		\toprule
		                                              & network                                                  & type & \textbf{mIoU}$\,\uparrow$ & Vegetation & Terrain & Vehicle & Object & Constr. & Road & Sign & Human & Sky \\

		\midrule
		\multirow{4}{*}{\rotatebox[origin=c]{90}{2D}} &
		\multirow{2}{*}{PP-LiteSeg \cite{peng_ppliteseg_2022}}
		                                              & \textit{category}
		                                              & \textbf{63.60}
		                                              & 85.62
		                                              & 85.84
		                                              & 64.10
		                                              & 44.83
		                                              & 72.08
		                                              & 53.84
		                                              & 1.19
		                                              & 67.21
		                                              & 97.67
		\\
		                                              &
		                                              & \textit{class}
		                                              & \textbf{43.83}
		                                              & 26.83
		                                              & 65.50
		                                              & 27.53
		                                              & 18.78
		                                              & 65.63
		                                              & 24.00
		                                              & 3.79
		                                              & 65.47
		                                              & 96.99
		\\
		                                              & \multirow{2}{*}{DDRNet~\cite{pan_ddrnet_2022}}
		                                              & \textit{category}
		                                              & \textbf{62.03}
		                                              & 85.03
		                                              & 85.85
		                                              & 59.81
		                                              & 45.09
		                                              & 67.18
		                                              & 54.02
		                                              & 1.73
		                                              & 61.79
		                                              & 97.76
		\\
		                                              &
		                                              & \textit{class}
		                                              & \textbf{47.28}
		                                              & 42.25
		                                              & 70.22
		                                              & 27.81
		                                              & 26.60
		                                              & 66.65
		                                              & 31.74
		                                              & 2.15
		                                              & 61.28
		                                              & 96.77
		\\
		\midrule
		\multirow{4}{*}{\rotatebox[origin=c]{90}{3D}}
		                                              & \multirow{2}{*}{PTv3~\cite{wu_ptv3_2024, pointcept2023}}
		                                              & \textit{category}
		                                              & \textbf{63.83}
		                                              & 73.96
		                                              & 34.12
		                                              & 28.68
		                                              & 59.21
		                                              & 76.10
		                                              & 70.87
		                                              & 85.79
		                                              & 81.94
		                                              & -
		\\
		                                              &
		                                              & \textit{class}
		                                              & \textbf{29.27}
		                                              & 34.00
		                                              & 23.39
		                                              & 32.54
		                                              & 29.15
		                                              & 49.31
		                                              & 23.34
		                                              & 40.99
		                                              & 30.74
		                                              & -
		\\
		                                              & \multirow{2}{*}{MSeg3D~\cite{jiale_mseg3d_2023}}
		                                              & \textit{category}
		                                              & \textbf{36.26}
		                                              & 51.55                                                                                                                                                                       
		                                              & 80.71                                                                                                                                                                       
		                                              & 42.53                                                                                                                                                                       
		                                              & 29.81                                                                                                                                                                       
		                                              & 19.26                                                                                                                                                                       
		                                              & 14.11                                                                                                                                                                       
		                                              & 32.33                                                                                                                                                                       
		                                              & 19.78                                                                                                                                                                       
		                                              & -
		\\
		                                              &
		                                              & \textit{class}
		                                              & \textbf{20.87}
		                                              & 27.52
		                                              & 27.78
		                                              & 30.67
		                                              & \phantom{0}1.61
		                                              & 41.99
		                                              & 40.38
		                                              & 17.86
		                                              & \phantom{0}0.00
		                                              & -
		\\
		\bottomrule
	\end{tabular}
	\label{tab:evaluation}
\end{table*}

\begin{table}[htpb!]
	\centering
	\caption{Fine-tuning performance of the GOOSE-Ex dataset. The number in brackets displays the IoU delta between the models trained only on GOOSE and after the fine-tuning.}
	\renewcommand{\arraystretch}{1.1}
	\begin{tabular}{*{8}{cllcc}}
		\toprule
		                                              & network                                                  & split & \makecell{\textbf{class} \\ \textbf{mIoU}$\,\uparrow$} & \makecell{\textbf{category} \\ \textbf{mIoU}$\,\uparrow$}\\

		\midrule
		\multirow{6}{*}{\rotatebox[origin=c]{90}{2D}} &
		\multirow{3}{*}{PP-LiteSeg \cite{peng_ppliteseg_2022}}
		                                              & All
		                                              & 43.83 (\textbf{+28.71})
		                                              & 63.60 (\textbf{+24.91})
		\\
		                                              &
		                                              & Alice
		                                              & 26.89 (\textbf{+24.75})
		                                              & 47.89 (\textbf{+22.31})
		\\
		                                              &
		                                              & Spot
		                                              & 47.82 (\textbf{+30.58})
		                                              & 64.71 (\textbf{+20.58})
		\\\cmidrule{2-5}
		                                              & \multirow{3}{*}{DDRNet~\cite{pan_ddrnet_2022}}
		                                              & All
		                                              & 47.28 (\textbf{+39.79})
		                                              & 62.03 (\textbf{+25.24})
		\\
		                                              &
		                                              & Alice
		                                              & 28.29 (\textbf{+25.39})
		                                              & 42.53 (\textbf{+18.72})
		\\
		                                              &
		                                              & Spot
		                                              & 49.63 (\textbf{+39.56})
		                                              & 65.08 (\textbf{+22.11})
		\\
		\midrule
		\multirow{6}{*}{\rotatebox[origin=c]{90}{3D}}
		                                              & \multirow{3}{*}{PTv3~\cite{wu_ptv3_2024, pointcept2023}}
		                                              & All
		                                              & 29.27 (\textbf{+14.26})
		                                              & 63.83 (\textbf{+32.96})
		\\
		                                              &
		                                              & Alice
		                                              & 17.18 (\textbf{+\phantom{0}7.50})
		                                              & 57.42 (\textbf{+29.13})
		\\
		                                              &
		                                              & Spot
		                                              & 28.65 (\textbf{+11.04})
		                                              & 70.71 (\textbf{+30.07})
		\\\cmidrule{2-5}
		                                              & \multirow{3}{*}{MSeg3D~\cite{jiale_mseg3d_2023}}
		                                              & All
		                                              & 20.87 (\textbf{+12.18})                                                                     
		                                              & 36.26 (\textbf{+22.18})                                                                     
		\\
		                                              &
		                                              & Alice
		                                              & 14.77 (\textbf{+\phantom{0}8.10})                                                           
		                                              & 34.23 (\textbf{+20.14})                                                                     
		\\
		                                              &
		                                              & Spot
		                                              & 27.91 (\textbf{+14.10})                                                                     
		                                              & 60.85 (\textbf{+31.03})                                                                     
		\\
		\bottomrule
	\end{tabular}
	\label{tab:finetuning_evaluation}
\end{table}

\subsection{Evaluation Metrics}
\label{subsec:evaluation-metrics}

The standard metric for evaluating the semantic segmentation on both images and point clouds is the mean Intersection over Union (mIoU). For each class $i$, it compares the prediction region with the ground truth region, resulting in
\begin{align}
	\text{IoU}_{i} = \frac{\sum_{I} \sum_{x,y}  \mathds{1}(P(x,y) == i \;\land\; GT(x,y) == i)}{\sum_{l} \sum_{x,y}  \mathds{1}(P(x,y) == i  \;\lor\; GT(x,y) == i)}
\end{align}
\noindent with $I$ being the image, $P(x,y)$ the predicted label, $GT(x,y)$ the ground truth and $\mathds{1}$ the indicator function. The IoU is accumulated over the entire test-set and averaged over all classes to yield the mIoU.
The IoU metric is known to be biased towards object instances and classes that cover large areas of the image~\cite{cordts_cityscapes_2016}, therefore fine-grained datasets with many classes as ours generally produce weaker mIoU results.

\subsection{Semantic Segmentation}
\label{subsec:semantic-segmentation}

In Table~\ref{tab:evaluation}, we provide averaged IoU values for different state-of-the-art 2D and 3D semantic segmentation models which were trained on the full set of classes \textit{(class)} as well as IoU values for models trained on the broader category labels \textit{(category)}.
\textit{PPLiteSeg}~\cite{peng_ppliteseg_2022} uses an encoder-decoder structure with a lightweight attention-based fusion model in the decoder to enable real-time semantic segmentation.
\textit{DDRNet}~\cite{pan_ddrnet_2022} is based on \textit{BiSeNet}~\cite{yu_bisenet_2018}, which uses a typical two-stream architecture and fuses both branches at different depths in the network.
For 3D segmentation, the recent Point Transformer V3~\cite{wu_ptv3_2024} (\textit{PTv3}) makes use of so-called Point Transformer layers on the point cloud input as its building blocks in a encoder-decoder architecture similar to \textit{U-Net}~\cite{ronneberger_unet_2015}.
We use the lidar-only variant of \textit{MSeg3D}~\cite{jiale_mseg3d_2023} that uses a voxel-based feature encoder with sparse convolutions for point-wise feature learning similar to Part-A\textsuperscript{2}~\cite{shi_SparseConvUNet_2021}.

We observe a good performance on labels with a high presence in the data (e.g. \textit{vegetation}, \textit{terrain}, \textit{sky}) and an expected lower performance on poorly represented classes. The results are very similar and comparable to those obtained in \cite{mortimer_goose_2024}, with the difference that the scenarios and platforms represented in the GOOSE-Ex data are more diverse than those of the original GOOSE dataset. When the category models are evaluated on the original GOOSE test split, a mIoU of \textit{55.75\%} and \textit{49.75\%} is obtained for PP-LiteSeg and DDRNet respectively, showing good generalization capabilities on all scenarios. A platform-specific comparison can be seen in Table~\ref{tab:finetuning_evaluation}. Here we observe a lower performance on the excavator platform, due to an unorthodox field of view that observes mostly ground pixels and points. The appearing vegetation classes are harder to distinguish, especially for the 3D cases, and simple classes like \textit{Sky} appear less often. As stated above, the \textit{class} results are quite low compared to other datasets due to several very rare classes that produce IoU values of zero. The impressive performance of the PTv3~\cite{wu_ptv3_2024} model on categories suggests that 3D segmentation models can provide valuable input for robust navigation solutions even in difficult unstructured environments.

\section{Downstream Applications}
\label{sec:downstream}

We verify the practicability of the GOOSE-Ex dataset on multiple downstream applications.

\subsection{Terrain Traversability Estimation}
\label{subsec:traversability}

In many cases, coarse-grained semantic segmentation is sufficient to interface a traversability analysis with a path planner. \textit{GANav}~\cite{ganav} proposes a group-wise attention mechanism to segment RGB images of unstructured environments into navigable regions. The attention mechanism and the corresponding group-wise attention loss help the transformer architecture to efficiently fuse multi-scale image features. The approach has originally been tested on the RUGD~\cite{wigness_rugd_2019} and the RELLIS-3D~\cite{jiang_rellis-3d_2021} datasets. We train GANav on the GOOSE and GOOSE-Ex datasets and test on all four datasets to verify the generalizability properties. We follow the original categorization into 6 semantic classes: \textit{smooth}, \textit{rough}, \textit{bumby}, \textit{forbidden}, \textit{obstacle}, \textit{background}.

\begin{table}[htpb!]
	\centering
	\caption{Results of GANav~\cite{ganav} trained and tested on different datasets to investigate their generalizability.}
	\begin{tabular}{lrrrr}
		\toprule
		     & \multicolumn{4}{c}{\textbf{mIoU}$\,\uparrow$} \\
		test & \makecell[l]{train on                         \\RUGD} & \makecell[l]{train on \\RELLIS-3D} & \makecell[l]{train on \\GOOSE} & \makecell[l]{fine-tune on\\ GOOSE-Ex} \\
		\midrule
		RUGD~\cite{wigness_rugd_2019}
		     & 89.08
		     & 15.38
		     & 21.59
		     & 29.35
		\\
		RELLIS-3D~\cite{jiang_rellis-3d_2021}
		     & 24.76
		     & 74.44
		     & 36.86
		     & 45.56
		\\
		GOOSE~\cite{mortimer_goose_2024}
		     & 17.74
		     & 17.87
		     & 37.99
		     & 41.36
		\\
		GOOSE-Ex
		     & 22.95
		     & 19.74
		     & 31.45
		     & 54.89
		\\
		\bottomrule
	\end{tabular}
	\label{tab:ganav}
\end{table}
\begin{figure}[ht]
	\vspace{-1em}
	\centering
	\includegraphics[width=0.99\columnwidth]{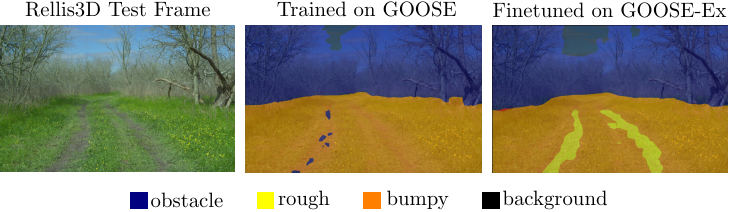}
	\caption{\label{fig:ganav_imgs} GANav~\cite{ganav}: Qualitative Segmentation on Unseen Dataset. Fine-tuning on GOOSE-Ex exhibits the best generalization to other datasets like RELLIS-3D~\cite{jiang_rellis-3d_2021}.}
\end{figure}
According to the GANav~\cite{ganav} segmentation results in Table~\ref{tab:ganav} and Figure~\ref{fig:ganav_imgs}, the GOOSE-Ex fine-tuning achieves the best generalization results across all four datasets. The high mIoU results on the RELLIS-3D and RUGD test sets trained on their respective train sets indicate a low inter-set variance compared to the GOOSE datasests. Also, the models have not been optimized for a high performance on the GOOSE datasets.

\subsection{Object Manipulation}
\label{subsec:manipulation}

\begin{figure}[htpb!]
	\centering
	\includegraphics[width=1.0\columnwidth]{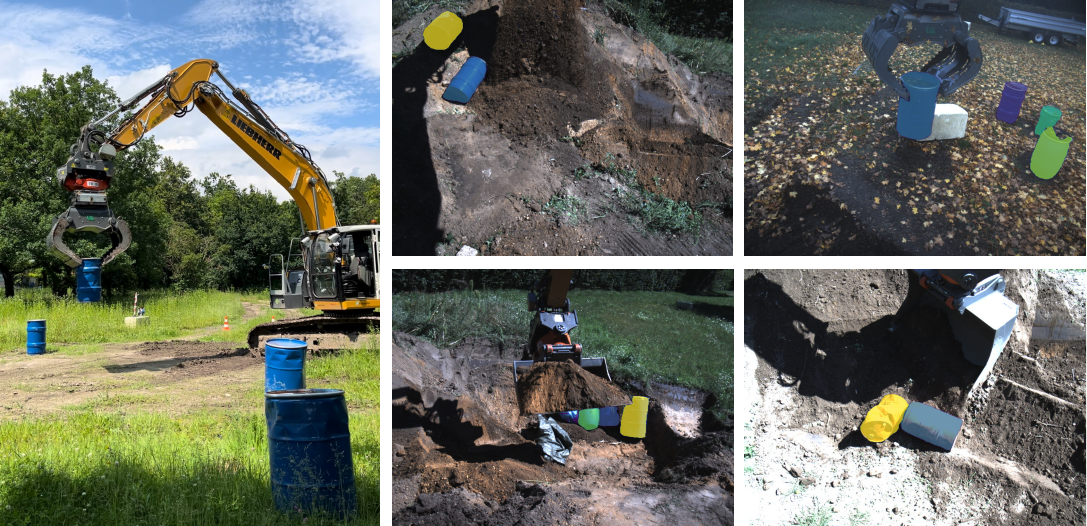}
	\caption{
		Excavator \textit{ALICE} grasping barrels (left) using panoptic segmentation masks (right) obtained from a Mask2Former~\cite{cheng2021mask2former} model trained on the GOOSE-Ex dataset.
	}
	\label{fig:instance_segmentation}
\end{figure}

Construction machines are often employed to manipulate heavy or hazardous objects. For these tasks, it is crucial to accurately identify and distinguish the target object from the surrounding environment. We trained \textit{Mask2Former}~\cite{cheng2021mask2former} as a panoptic segmentation approach on the GOOSE-Ex datasets to exploit the instance labels of many objects in the dataset. Specifically, we trained a model for panoptic segmentation of barrels, some qualitative results are shown in Figure~\ref{fig:instance_segmentation}. After a fusion with the depth perception, this allows the excavator to perform accurate pose estimation and subsequent grasping of barrels in complex environments.

\subsection{SLAM Benchmark and Semantic Scene Completion}
\label{subsec:slam}

\begin{figure}[htpb!]
	\centering
	\subfloat[Semantic Scene GT for SSC Task on IOSB Campus]{%
		\includegraphics[height=3cm]{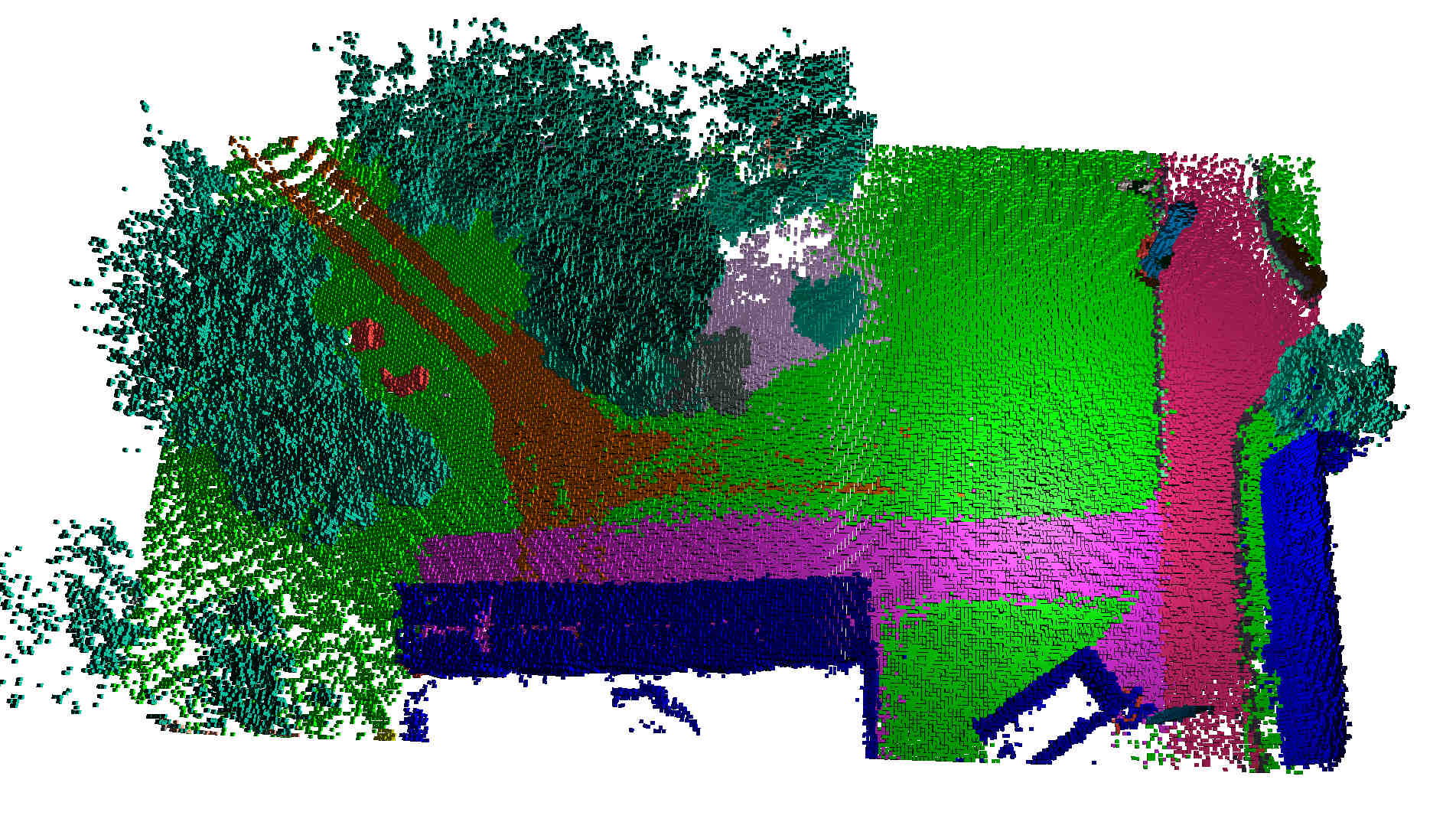}%
		\includegraphics[height=3cm]{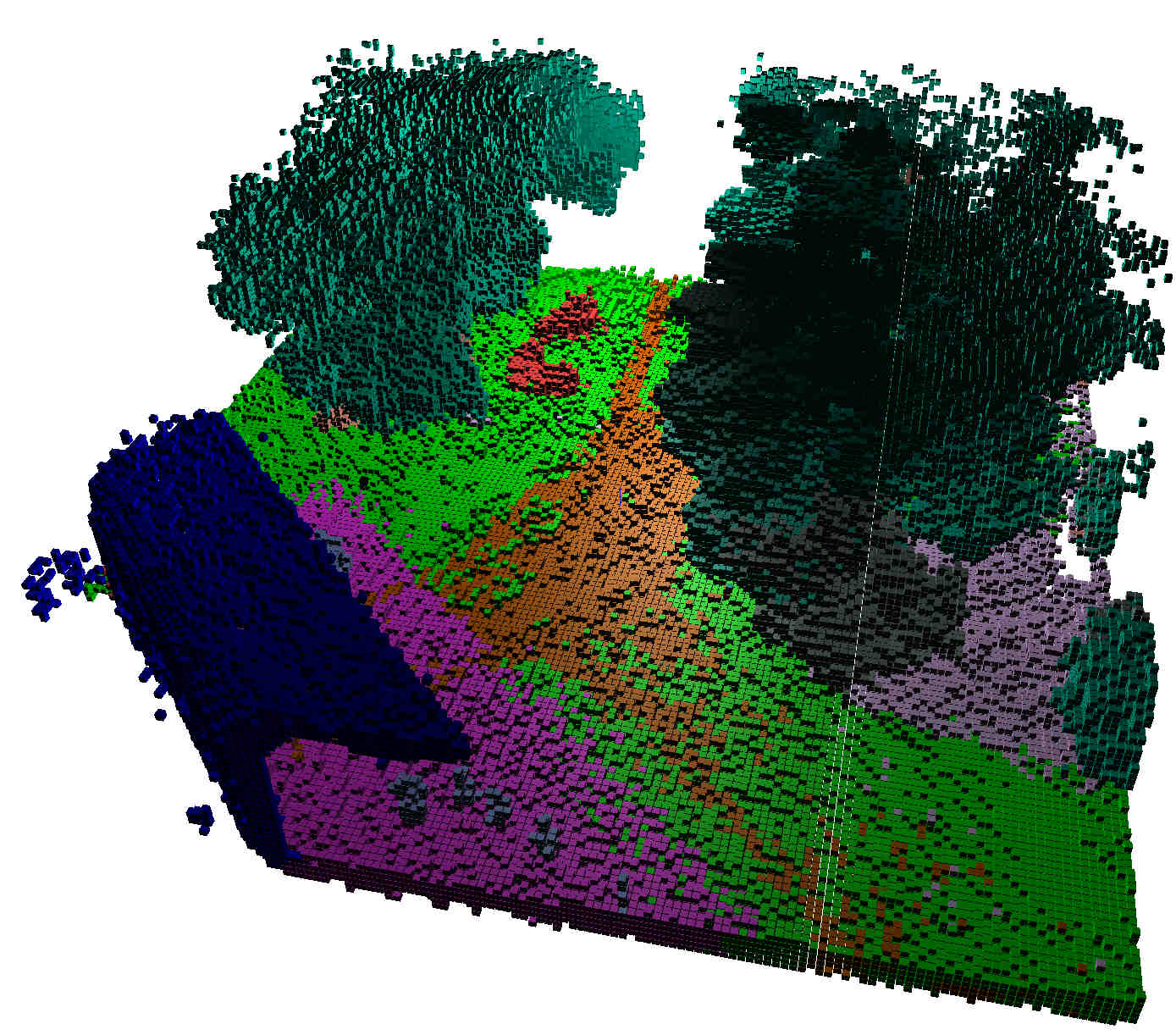}%
		\label{fig:semanticscene}%
	}\\
	\subfloat[IOSB Campus]{%
		\includegraphics[height=2.2cm]{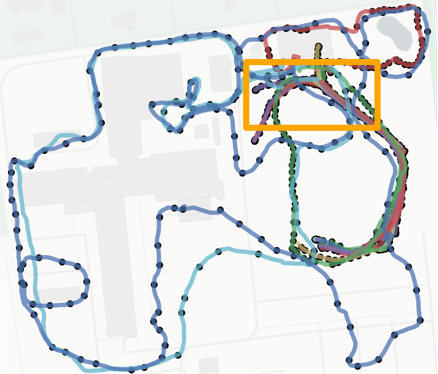}%
		\label{fig:map:a}%
	} \,
	\subfloat[Landfill]{%
		\includegraphics[height=2.2cm]{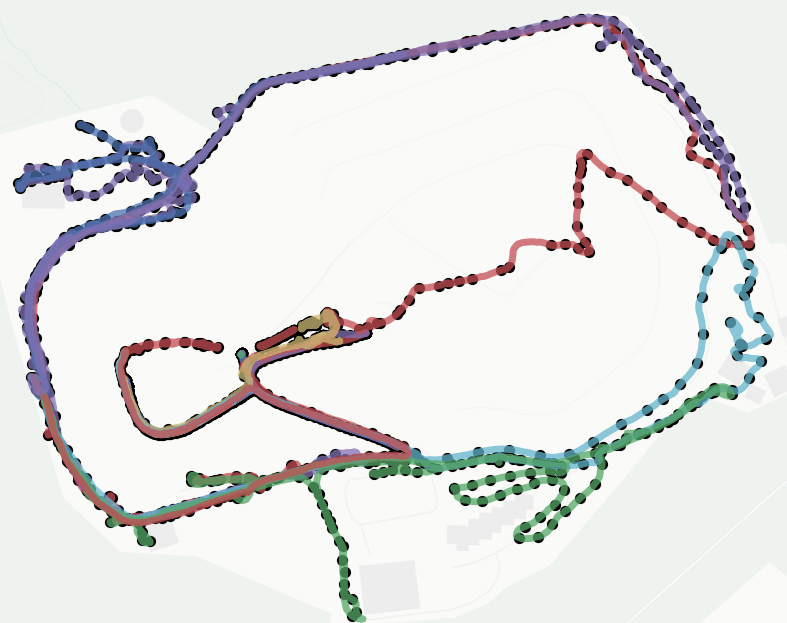}%
		\label{fig:map:b}%
	} \,
	\subfloat[Construction Site]{%
		\includegraphics[height=2.2cm]{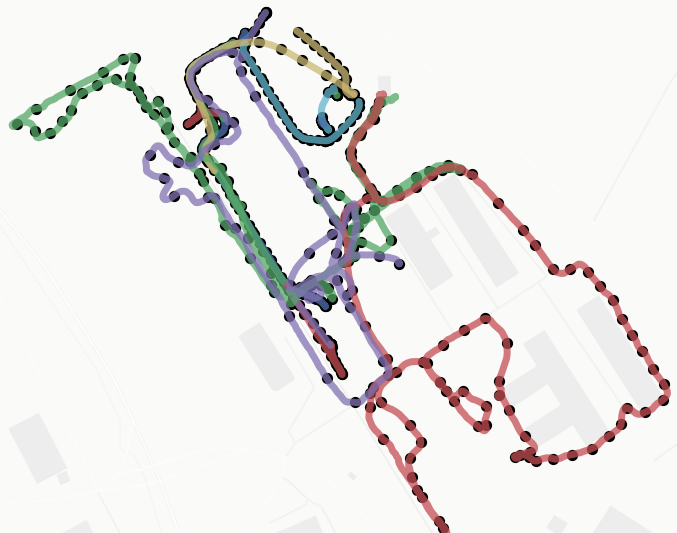}%
		\label{fig:map:c}%
	}
	\caption{
		A selection of sequences that can be used as testbed for (semantic)-SLAM approaches. Dots $\bullet$ indicate an annotated frame. The highlighted area \includegraphics[height=0.7em]{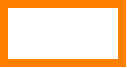} in (b) corresponds to the reconstructed semantic scene displayed in (a). Background maps: \textcopyright OpenStreetMap contributors
	}
	\label{fig:trajectories}
	\vspace{-1em}
\end{figure}

Beyond the scope of semantic segmentation, the GOOSE-Ex dataset can be used as an odometry or SLAM test set for several tasks: Single-robot odometry estimation, loop-closure detection based on images or point clouds, loop-closure detection based on semantics, or even multi-robot SLAM. Figures~\ref{fig:map:a}-\ref{fig:map:c} provide an overview of sequences of selected scenarios that may prove useful for these tasks. All sequences come with a GNSS-based ground truth estimate.

We have post-processed several groups of sequences with a SLAM approach~\cite{Emter2018} based on GTSAM~\cite{gtsam} to achieve a high annotation density for different regions. By fusing multiple frames we can generate ground truth semantic maps that can be used to tackle the task of Semantic Scene Completion (SSC)~\cite{ssc-survey} in unstructured environments. The goal of the SSC task is to predict both geometry and semantics for a specified target volume, given a single LiDAR scan as input. We use the processing pipeline of SemanticKITTI~\cite{behley_semantickitti_2019} to generate voxel maps with a voxel resolution of $\SI{0.2}{\meter}$. Figure~\ref{fig:semanticscene} shows an example region. We plan to release this as standalone SSC dataset in the future, which would allow fine-tuning SSC approaches for unstructured environments.

\section{Conclusion}
\label{sec:conclusion}
We present the GOOSE-Ex dataset for semantic segmentation in unstructured environments across domains and platforms. In future work, we want to explore approaches that benefit from the multimodality of the dataset and extend the research scope from semantic segmentation to other tasks such as semantic SLAM.
\\\\
\noindent\textbf{Acknowledgements}
We thank the Autonomous Robotic Systems group at Fraunhofer IOSB for their support during several data collection trips. We also thank Roman Abayev and Anselm von Gladiß for maintaining the GOOSE DB for querying sequences within the GOOSE datasets.

\balance
\bibliographystyle{IEEEtran}
\bibliography{bib/IEEEfull,bib/additional_full,bib/literature}

\end{document}